%% file: paper.tex
\documentclass[aoas,preprint]{imsart}

\usepackage{xcolor} 
\RequirePackage[OT1]{fontenc}
\RequirePackage{amsthm,amsmath}
\RequirePackage[numbers]{natbib}
\RequirePackage[colorlinks,citecolor=blue,urlcolor=blue]{hyperref}
\usepackage{url}
\usepackage{amssymb}
\usepackage{graphicx}
\usepackage[export]{adjustbox}
\usepackage{caption}
\usepackage{subcaption}
\startlocaldefs
\numberwithin{equation}{section}
\theoremstyle{plain}

\endlocaldefs

\newif\ifgray
\graytrue
\grayfalse
\IfFileExists{BackgroundColor.tex}{\input{BackgroundColor}}{}
\ifgray
\pagecolor{black!30!white}
\fi

\begin{document}

\newcommand{ \AddInfo}{Additional Knowledge}
\newcommand{ \Addinfo}{Additional knowledge}
\newcommand{ \addinfo}{additional knowledge}
\newcommand{ \ourmethod}{SI}

\begin{frontmatter}
\title{Integrating  \addinfo\ Into\\
Estimation of Graphical Models}
\runtitle{Graphical Modeling With \AddInfo}

\begin{aug}
\author{\fnms{Yunqi} \snm{Bu}\ead[label=e1,email]{yunqibu@uw.edu}}\and\author{\fnms{Johannes} \snm{Lederer}\ead[label=e2,email]{ledererj@uw.edu}
\ead[label=u1,url]{http://www.johanneslederer.com}}

\runauthor{Y. Bu and J. Lederer}

\affiliation{University of Washington
}

\address{Yunqi Bu\\
Department of Biostatistics\\ 
University of Washington\\
Seattle, WA 98195, U.S.A.\\
\printead{e1}
\phantom{E-mail:\ }\printead*{}}

\address{Johannes Lederer\\
Department of Statistics,\\
\ \ \ Biostatistics\\
University of Washington\\
Seattle, WA 98195, U.S.A.\\
\printead{e2}\\
\printead{u1}}
\end{aug}

\begin{abstract}
In applications of graphical models, we typically have more information than just the samples themselves. A prime example is the estimation of brain connectivity networks based on fMRI data, where in addition to the samples themselves, the spatial positions of the measurements are readily available. With particular regard for this application, we are thus interested in ways to incorporate additional knowledge most effectively into graph estimation. Our approach to this is to make neighborhood selection receptive to additional knowledge by strengthening the role of the tuning parameters. We demonstrate that this concept (i)~can improve reproducibility, (ii)~is computationally convenient and efficient, and (iii)~carries a lucid Bayesian interpretation. We specifically show that the approach provides effective estimations of brain connectivity graphs from fMRI data. However, providing a general scheme for the inclusion of \addinfo, our concept is  expected to have applications in a wide range of domains.
\end{abstract}



\end{frontmatter}

 \newcommand{\graphset}{\mathcal G}
 \newcommand{\nodeset}{\mathcal V}
 \newcommand{\edgeset}{{\mathcal E}}
 \newcommand{\covariancematrix}{\Sigma}
 \newcommand{\precisionmatrix}{{\Sigma^{\text -1}}}
 \newcommand{\adjacencymatrix}{A}
 \newcommand{\numberofedges}{k}
 \newcommand{\edge}{e}
\newcommand{\rnp}{ \mathbb R^{n\times p}}
\newcommand{\rpp}{ \mathbb R^{p\times p}}
\newcommand{\normone}[1]{\ensuremath{|\!|#1|\!|_1}}
\newcommand{\normtwo}[1]{\ensuremath{|\!|#1|\!|_2}}
\newcommand{\norm}[1]{\ensuremath{|\!|#1|\!|}}
\newcommand{\so}{\vspace{1mm}}
\newcommand{\st}{\hline\vspace{-1mm}\\}
\newcommand{\str}{~\vspace{4mm}\\}  
\newcommand{\nameGL}{GLASSO}
\newcommand{\nameNA}{MB-and}
\newcommand{\nameNO}{MB-or}
\newcommand{\nameJJO}{SI-or}
\newcommand{\nameJJA}{SI-and}
\newcommand{\nameJJ}{SI}
\newcommand{\amin}{\textcolor{black}{0.2}}
\newcommand{\amax}{\textcolor{black}{1}}
\newcommand{\conditionvalue}{100}


\section{Introduction} 
\input{Intro}


\section{Method}
	\label{method}
We start with a brief review of Gaussian graphical models. We then demonstrate the need for \addinfo\ in graph estimation with a simulation study that adopts the dimensions of the brain connectivity application. We are then ready to introduce and discuss our approach for amending the estimation process with \addinfo. We finally confirm by using a simulation that our method can improve graph estimation.

\input{ReviewGGM}

\input{NeedSI}

\input{MBSI}

 	\label{ourmethod}
\input{ROC.tex}


\section{Brain connectivity application}
	\label{App}
The main motivation for the general framework in Section~\ref{method} is brain connectivity networks. Our biological rationale for these networks is that direct connections are more likely between close regions than between distant regions. The novel concept we proposed is ideal for incorporating this notion. First, graphical models in general distinguish between direct connections (conditional dependence) and indirect connections (marginal dependence). Second, the mentioned rationale that the distance between two nodes influences the likelihood of them being directly connected can be reflected naturally by how the link function~$f$ incorporates the additional distance information~$D$. Importantly, we do not generally exclude the direct or indirect long-range connections. This means that two distant nodes can very well be connected, but more likely, this connection will be indirect. While we work with specific data about Alzheimer's disease in the following, our main goal is to show that our method provides a sensible approach to brain connectivity in general.
\input{bgdata}

\input{specify}

\input{stable} 
\input{connect}

	\label{connect}
	

\section{Discussion and further research}
	\label{diss}
\input{diss}

\appendix

\section*{Acknowledgements} We thank Andrew Zhou for the inspiring discussions and Rosemary Adams for contributions to the visualizations.  We also thank Noah Simon, Sam Koelle, and Mengjie Pan for helpful suggestions.

\end{document}

%% file: Intro.tex
Brain connectivity networks derived from fMRI data are considered a prime gateway to understanding  cognitive diseases. Along with  other fields, brain research has thus boosted the interest in statistical methodology for uncovering dependence networks. A standard framework for dependence networks is Gaussian graphical models~\cite{graph}. Gaussian graphical models have become particularly popular after the development of methods and algorithms that can handle large and high-dimensional data.

Two widely-used approaches to Gaussian graphical models are neighborhood selection and graphical lasso. Neighborhood selection aims at graph reconstruction by aggregating local estimates~\cite{MB}. Graphical lasso, on the other hand, is based on a global objective function~\cite{banerjee,GL,gl2}. Both approaches are now accompanied by a bulk of literature on theory and computation; we refer to~\cite{high, learn} and references therein. However, standard estimators do not take into account \addinfo\  that is often available in practice, such as scientific rationales, experimental arrangements, or insights from previous studies.  A natural question is thus how graph estimation can be refined when \addinfo\ is available.

The purpose of this paper is to study this question. Our main inspiration and application are brain connectivity networks. In particular, we are interested in estimating brain connectivity networks by analyzing resting-state functional magnetic resonance imaging (fMRI) data that describe the levels of co-activation between brain regions~\cite{fMRI} as measured by changes in blood flow~\cite{blood flow}. Brain regions have spatial coordinates, so in addition to the samples, there is information in terms of pairwise distances between regions. Our goal is to leverage this \addinfo\ for effective graph estimation. 

Our main idea for this is to strengthen the role of tuning parameters. Commonly, tuning parameters are considered an inconvenience, because they need to be calibrated for each data set specifically. We instead think of this adaptability as an asset that can make tuning parameters a potent instrument for funneling external information into the estimation process. More specifically, for our goal of brain network estimation, we use tuning parameters to make neighborhood selection  receptive to \addinfo. We first show numerically that without such an integration, the sample sizes needed for accurate graph recovery are surprisingly large. We then show that adopting our notion can lead to three main features.

{\it Reproducibility:} Reproducibility is a major principle in the sciences and has become again a heated topic in recent literature~\cite{NIH reproduce, Biost reproduce, imp reproduce}. We thus adopt a notion of reproducibility for our data application to show the competitiveness of our approach.

{\it Implementation:} We point out that our approach is amenable to straightforward and efficient implementations based on existing software. In particular, our approach preserves the general forms of the standard penalties and does not introduce any additional penalty terms.

{\it Interpretation:} We demonstrate that our approach has a lucid Bayesian interpretation.

The remainder of the paper is structured as follows. In Section~\ref{method}, we introduce and motivate our general concept. In Section~\ref{App}, we then tailor this concept to the estimation of brain connectivity networks based on fMRI data. In Section~\ref{diss}, we conclude with a discussion.

%% file: ReviewGGM.tex
\subsection{A brief review of Gaussian graphical models}
Gaussian graphical models assume samples from a centered $p$-dimensional normal distribution $\mathcal N_p (0,\Sigma)$ with a symmetric, positive definite covariance matrix~$\Sigma$. The distribution is then complemented with an associated undirected graph $\graphset=(\nodeset,\edgeset)$ that has node set  $\nodeset=\{1,\dots,p\}$ and edge set $\edgeset=\{(i,j)\in\nodeset\times \nodeset:(\precisionmatrix)_{ij}\neq 0\}$.

The crux of Gaussian graphical models is that the conditional and marginal dependence structures of the samples are concisely captured by the edge set~$\edgeset$. Indeed, the Hammersley-Clifford theorem~\cite{Hammersley1, Hammersley2} states that the $i$th and $j$th coordinate of a sample from $\mathcal N_p(0,\Sigma)$ are conditionally independent given all other coordinates if and only if $(\Sigma^{-1})_{ij}=0$, that is, $(i,j)\notin\edgeset$. Moreover, the $i$th and $j$th coordinate are independent if and only if one cannot construct a chain of the form  $(i,k_1),(k_1,k_2),\dots,(k_l,j)$ by using only elements in~$\edgeset$. Our goal is consequently to uncover the edge set $\edgeset$ from data. For this aim, we develop estimators $\widehat\edgeset\equiv\widehat\edgeset(X)$ of $\edgeset$ based on independent  identically distributed Gaussian samples $X_1,\dots,X_n\in\mathbb R^p$ summarized in the data matrix $X=(X_1,\dots,X_n)^\top\in\mathbb R^{n\times p}$.

%% file: NeedSI.tex
\subsection{The need for \addinfo}\label{sec:needsi}
We now demonstrate that even with optimal tuning and large sample sizes, standard methods for Gaussian graphical modeling can fail to provide accurate graph recovery. For this, we conduct a simulation study comparing four of the most popular methods: thresholding the partial correlation matrix (THR); neighborhood selection with the ``or-rule" (MB(or)); neighborhood selection with the ``and-rule'' (MB(and)); and graphical lasso (GLASSO).
  
We simulate data with the aim of mimicking settings encountered in practice. For this, the number of nodes is set to $p=116$, which equals the number of brain regions in the data considered in Section~\ref{App}. Then, a standard preferential attachment algorithm~\cite{scale free} is used to construct $115$ edges between these nodes. The resulting edge set~$\edgeset$ determines which off-diagonal entries of the inverse covariance matrix~$\precisionmatrix$ are non-zero. The values of these entries are then independently sampled uniformly at random from $[-1, -0.2]\cup[0.2, 1]$. Diagonal entries of~$\precisionmatrix$ are set to a common value such that the  condition number equals $100$.  With~$\precisionmatrix$ constructed this way, vectors are independently sampled from the Gaussian distribution~$\mathcal N_p (0,\Sigma)$ and summarized in data sets with sample sizes $n \in \{$50, 100, 200, 400, 600, 800, 1000, 1200, 1400, 1600$\}$.

The accuracy of graph recovery is assessed via Hamming distance. The Hamming distance between the estimated edge sets  $\tilde\edgeset$ and the true edge set~$\edgeset$ is defined by $d_{\operatorname{H}}(\tilde \edgeset, \edgeset):=|\{(i,j):(i,j)\in\tilde\edgeset,(i,j)\not\in\edgeset\}\cup \{(i,j):(i,j)\not\in\tilde\edgeset,(i,j)\in\edgeset\}|$. Larger Hamming distances indicate less accurate estimation. The tuning parameters of all methods are calibrated to minimal Hamming distance, noting that the true graphs are known in simulations. This ``oracle'' tuning  allows us to study the maximal potential of the methods' accuracies.

\begin{figure}[h!] 
\includegraphics[width=.97\textwidth]{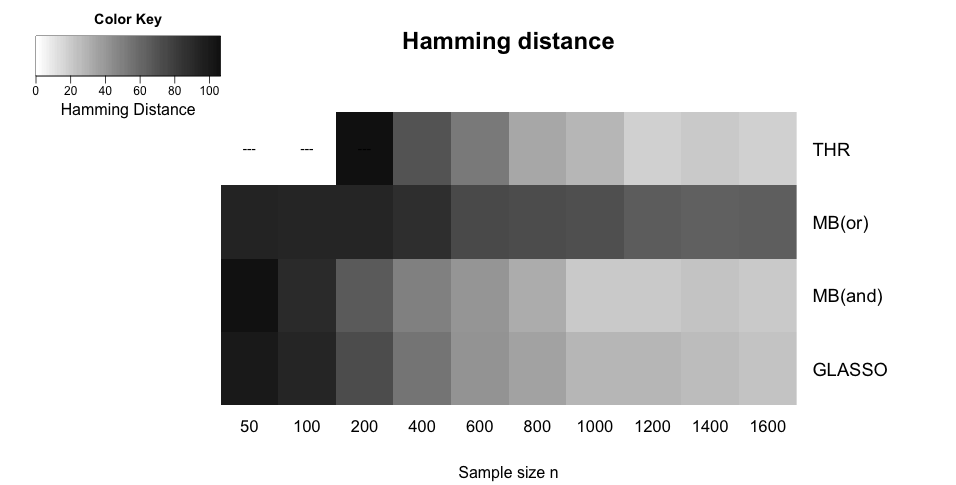} 
\caption[]{Graph estimation with optimally calibrated standard methods becomes more accurate as the sample size $n$ increases. However, even with  $n=1600$, which is much larger than the number of covariates $p=116$, the Hamming distances are $32$ or above. This means that graph estimation with standard methods can  still be highly inaccurate even when the sample sizes are very large.}
\label{simulation}
\end{figure}

Figure~\ref{simulation} contains the heat-map for accuracies averaged over $20$ repetitions. For $n=1600$, THR is the most accurate approach among all four methods, having average Hamming distance 32. Observe, however, that $32$ wrongly assigned edges still mark a poor performance given that there are only $115$ true edges in total. For smaller sample sizes, the accuracies of the methods decline even further (note that in particular, THR requires $p\leq n$ and thus can not be applied in the $n\in\{50,100\}$ regime as indicated by dashed lines in the figure). In view of real data being commonly high-dimensional, where $p$ is of the same order as $n$ - or even larger, these observations thus provide substantial motivation for the inclusion of \addinfo.

%% file: MBSI.tex
\subsection{Neighborhood selection with \addinfo} We now introduce our scheme to incorporate \addinfo\ into graph recovery. The resulting method can be computed efficiently with standard software packages and has a direct Bayesian interpretation.

A basis particularly suited for our approach is neighborhood selection. To recapitulate, the main idea of neighborhood selection is that graph recovery can be established through a sequence of regressions. The corresponding regression parameter $\beta^j\in\mathbb R^p$ for a given node $j$ determines the edges between node $j$ and the other nodes: $(\beta^j)_i\neq 0~\Leftrightarrow (i,j)\in\edgeset$. With this in mind, the standard estimators are of the form
\begin{equation}
\hat\beta^{j,r^j}\in\underset{\substack{\beta\in \mathbb R^p\\\beta_j=0}}
{\operatorname{arg min}}
\Big\{\frac{1}{2}\normtwo{X^j-X\beta}^2+r^j \norm{\beta}\Big\}\,,~~~~(\,j\in\{1,\dots,p\}\,)\,.
 \label{general mb}
 \end{equation} 
Here, the vector $X^j\in\mathbb R^n$ denotes the $j$th column of the data matrix~$X,$ the positive numbers $r^1,\dots,r^p\in [0,\infty)$ are tuning parameters, and $\norm{\cdot}$ is a norm (or more generally, a convex, positive function, such as the Ridge penalty). A standard example is neighborhood selection with the lasso, where $\norm{\cdot}$ is set to the $\ell_1$-norm. Adopting the ``and-rule", the edge set~$\edgeset$ is finally estimated by $\hat{\edgeset}=\big\{(i,j)\ |\ (\hat\beta^{i,r^i})_j\neq0 \text{~and~}(\hat\beta^{j,r^j})_i\neq 0\}$.

The choice of MB(and) as the basis is motivated by the simulation results in Section~\ref{sec:needsi} and the fact that there is a myriad of existing software packages for solving problems of the form~\eqref{general mb}. In principle, however, one could apply the following recipe also to MB(or) and GLASSO.

Our proposal is now to incorporate \addinfo\ by ``upgrading'' the univariate tuning parameters $r^1,\dots,r^p$ to vectors. For this, we assume \addinfo\ in the form of a matrix $D\in\mathbb R^{p\times p}.$ For example, $D_{ij}$ could be the Euclidean distance between nodes $i$ and $j$ in cases where the nodes correspond to brain regions, countries, galaxies, etc. As another example, $D_{ij}$ could be the (conditional) correlation between nodes $i$ and~$j$ estimated on the present data set with an initial estimator (specializing our method to adaptive lasso-type approaches~\cite{adpt lasso}, for example) or more interestingly, with the same or different estimators on other data sources. Next, to incorporate $D$ into neighborhood selection, we transform the one-dimensional tuning parameters $r^j\in[0,\infty)$ into multi-dimensional tuning parameters  $\mathbf{r}^j\in [0,\infty)^p$. We then enrich these tuning parameters with \addinfo\ by setting $(\mathbf{r}^j)_i=\bar r^j\cdot f(D_{ij}),$ where $f:\mathbb R\to\mathbb R$ is a link function and $\bar r^j$ is an overall tuning parameter  for the $j$th regression. As customary in high-dimensional statistics, the free parameter~$\bar r^j$ balances the weights of the data ($X$ in our case) and the structural assumptions (captured by $\norm{\cdot}$ and $D$ in our case). The above node-wise regression~\eqref{general mb} is then generalized to
\begin{equation}
\hat\beta^{j, \mathbf{r}^j}\in\underset{\substack{\beta\in \mathbb R^p\\ \beta_j=0}}
{\operatorname{arg min}}\Big\{\frac{1}{2}\normtwo{X^j-X\beta}^2+ \norm{\mathbf{r}^j \circ \ \beta }\Big\}\,,~~~~(\,j\in\{1,\dots,p\}\,)\,,
 \label{lassos} 
 \end{equation}
where the circle $\circ$ indicates element-wise multiplication.

Let us discuss three practical and methodological aspects of~\eqref{lassos}. The first important observation is that while there are now $p$ ``tuning'' parameters per regression, there is still only {\it one} free parameter (namely $\bar r^j$) that requires calibration. This is highly desirable: for one parameter, efficient calibration schemes are known~\cite{cv,ledererAVI,tune select}; in contrast, the calibration of multiple free parameters, which would appear when adding additional penalty terms instead of adopting our approach, remains a major challenge in both theory and computations. Second, for the link function $f$, there are often natural choices; we refer to the application section and the discussion. Third, note that our approach retains the ``flavor'' of the original penalty. This means that our concept maintains the general properties of the penalty, such as the sparsity generating effect of $\ell_1$-penalties~\cite{lasso}, and it means that the practical implementation of~\eqref{lassos} can be based on standard software packages, such as \texttt{glmnet}~\cite{glmnet} in the case of $\ell_1$-penalization.

Finally, our approach has a direct Bayesian interpretation. For a given index~$j$,  we consider the hierarchical Bayesian model 
\begin{equation}
X^j\, |\, \beta, \sigma \sim \mathcal N(X\beta\,, \sigma^2\operatorname{I}_{n\times n})\,,
 \label{Bayes y}
 \end{equation}
\begin{equation}
Pr(\beta_i\, |\, \sigma, (\mathbf{r}^j)_i) =  \frac{(\mathbf{r}^j)_i}{2\sigma^2}e^{- \frac{(\mathbf{r}^j)_i}{\sigma^2}|\beta_i|}\,,~~~~(\,i\in\{1,\dots,p\}\,)\,.
 \label{Bayes beta}
 \end{equation}
 This model is a generalization of the model considered in the seminal paper on the Bayesian lasso~\cite{Bayes lasso}. The parameters $(\mathbf{r}^j)_1,\dots,(\mathbf{r}^j)_p$ in our general model are hyperparameters  that specify the shape of the prior distribution of each of the regression coefficients~$(\beta^j)_i$. The negative log-posterior distribution of $\beta$ is now given by
 \begin{equation*}
\begin{split}
- \text{log}\, \mathbb P(\beta\, |\, X^j,\sigma, \mathbf{r}^j) & = \frac{1}{\sigma^2} \Big(\frac{1}{2}\normtwo{X^j- X\beta}^2+ \sum_{i=1}^p (\mathbf{r}^j)_i |\beta_i|\Big) +c\,,
\end{split}
 \label{neg2}
 \end{equation*}  
where $c$ is a term independent of $\beta$.  The mode of this distribution is
 \begin{equation*}
\hat\beta^{j,\mathbf{r}^j}\in\underset{\substack{\beta\in \mathbb R^p\\\beta_j=0}}
{\operatorname{arg min}} \Big\{ \frac{1}{2}\normtwo{X^j-X\beta}^2+ \sum_{i=1}^p
 (\mathbf{r}^j)_i |\beta_i| \Big\}\,,~~~~(\,j\in\{1,\dots,p\}\,)\,,
 \label{lasso}
 \end{equation*} 
which equals the estimator yielded by our approach when the penalty in~(\ref{lassos}) is set to the $\ell_1$-norm. Similarly, replacing the double-exponential distribution in~\eqref{Bayes beta} with a Gaussian distribution, the posterior mode equals the estimators yielded by our approach when the penalty in~(\ref{lassos}) is set to the $\ell_2$-norm.

This analysis shows that the tuning parameters relate our frequentist and Bayesian notions about the \addinfo. In our frequentist estimator, the larger $(\mathbf{r}^j)_i$, the more likely the edge $(i,j)$ is excluded from the estimate. In our Bayesian view, the larger $(\mathbf{r}^j)_i$, the more the assumed distribution of $(\beta^j)_i$ is concentrated around zero. Thus, funneling the \addinfo\ into the tuning parameters of the frequentist estimator can be viewed as transforming the original priors, which are the same all across the coefficient vector, into informed priors tailored to each coordinate.

%% file: ROC.tex
\subsection{Simulation} 
\begin{figure}
\centering
\begin{subfigure}{.4\textwidth}
  \centering
  \includegraphics[width=0.9\linewidth]{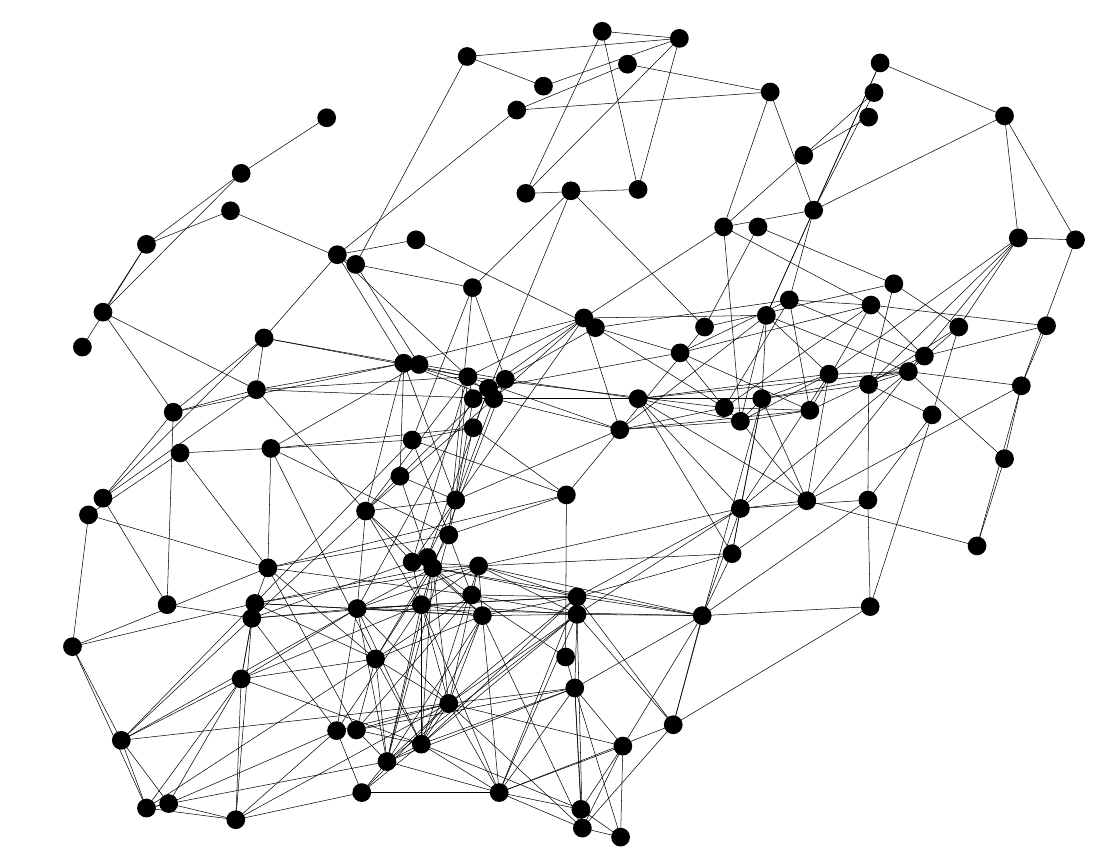}
\end{subfigure}%
\begin{subfigure}{.6\textwidth}
  \centering
  \includegraphics[width=1.0\linewidth]{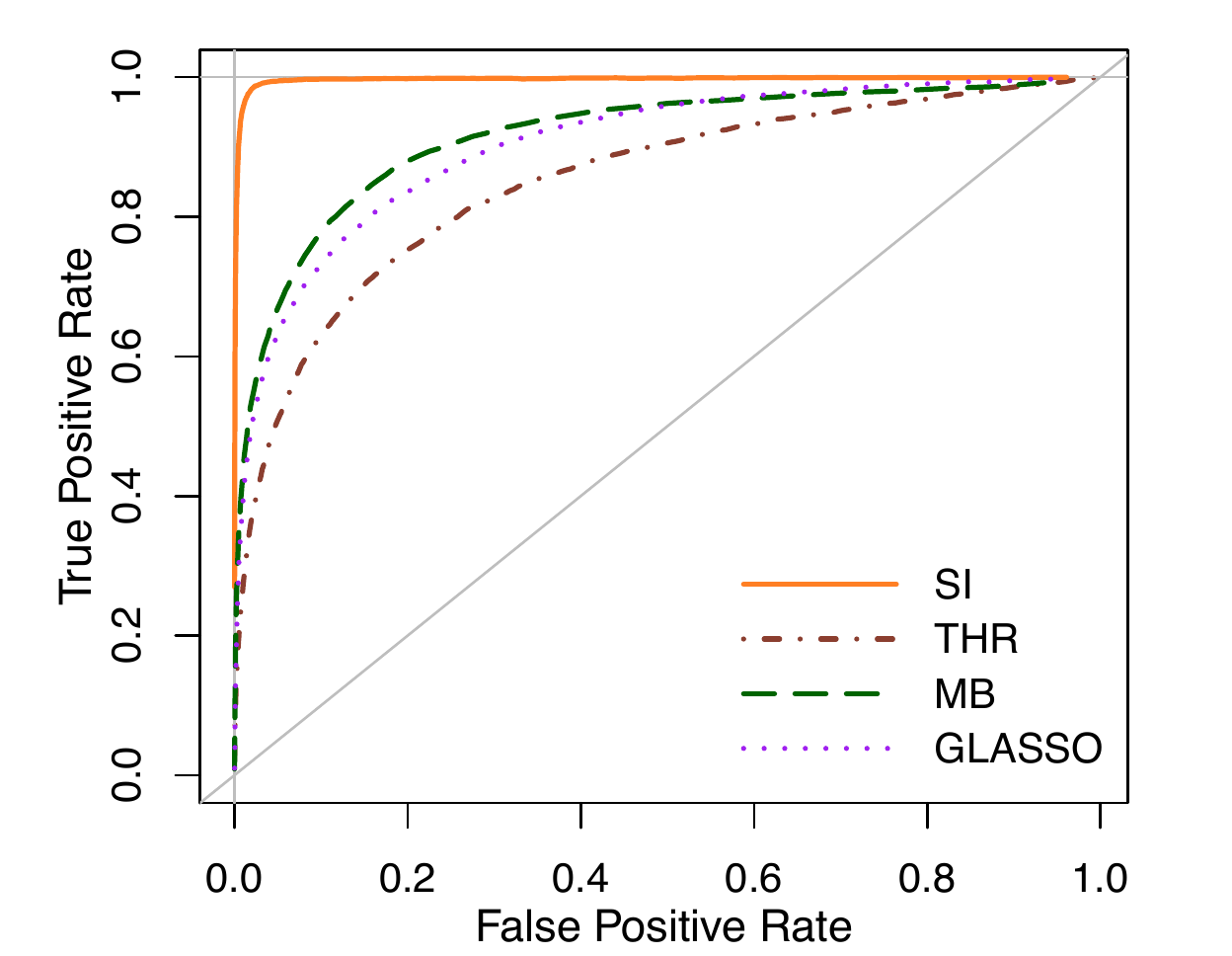}
\end{subfigure}
\caption{Left panel: One example of the $20$ simulated graphs. Right panel: ROC curves demonstrating that \ourmethod\ can outperform standard methods when \addinfo\ is available.}
\label{ROC}
\end{figure}

We now confirm by using a simulation that our approach can harvest additional knowledge to improve graph estimation. To generate data, we imitate the brain connectivity application in Section~\ref{App}: the number of nodes $p=116$ corresponds to the number of brain regions; the number of samples $n=210$ corresponds to the number of fMRI scans per subject; the \addinfo\ $D_{ij}$ is the Euclidian distance between brain regions $i$ and $j$. The edge set is constructed based on independent Bernoulli($p_{ij}$) distributions, $p_{ij}=\text{inv.logit}(10-D_{ij}/3)$, such that an edge $(i,j)$ is included if the corresponding Bernoulli outcome is one. The form of the distributions captures our rationale that direct connections are predominately between close regions. An anatomical map of a graph generated by the preceding scheme is displayed in the left panel of Figure~\ref{ROC}. Next, the non-zero entries in the inverse covariance matrix~$\precisionmatrix$ as specified by the edge set are set to 0.3. The diagonal entries are set to $0.2+0.3\cdot\sigma_{\min},$ where $\sigma_{\min}$ is the minimal singular value of the adjacency matrix. This construction ensures that $\precisionmatrix$ has full rank. Finally, $n$ i.i.d. samples are generated from~$\mathcal N_p (0,\Sigma)$.

We now run our approach against standard methods for graph estimation. The specification of our estimator, referred to as \ourmethod\ in the following, is the same as in the application section, namely neighborhood selection based on~\eqref{lassos} with $\ell_1$-norm penalty and link function~$f(x)=x^3$; see Section~\ref{specification} for further comments. The standard methods competing are THR, MB, and GLASSO. Motivated by the simulation results in Section~\ref{sec:needsi}, we adopt the ``and-rule'' for SI and MB. A total of $20$ data sets is generated, on which each method is run as a function of its free tuning parameter. The average ROC curves for graph recovery are plotted in the right panel of Figure~\ref{ROC}.  We find that the ROC curve of \ourmethod\ dominates the other curves, demonstrating that \ourmethod\ can improve graph recovery when \addinfo\ is available. 

We finally note that simulations alone cannot provide definite evidence in favor of a method, because data generating processes always reflect preconceived notions about the biological truth. We thus complement the simulation with a reproducibility study on real data in the following section. Together, the results provide us with sufficient confidence to argue in favor of \ourmethod.

%% file: bgdata.tex
\subsection{Background and data set} Functional Magnetic Resonance Imaging (fMRI) is a promising gateway to the understanding of the human brain~\cite{fMRI}. Our specific goal is to use brain activity records from fMRI to infer co-activation networks among brain regions. For this, we rely on data collected from outpatients at the Neurology Department of the Beijing Hospital from November 2013 through December 2015. The data set comprises~$37$ subjects: $22$ patients with Alzheimer's disease (AD), $5$ patients with mild cognitive impairment (MCI), and  $10$~patients with normal cognition (NC). Each subject's  data contains $n=210$ consecutive scans of the entire brain. Each of the $p=116$ variables is an average intensity over all voxels in an anatomical volume of interest defined by Automated Anatomical Labeling~\cite{116}. An autoregressive integrated moving average model~\cite{time1, time2} is applied to account for autocorrelation.

%% file: specify.tex
\subsection{Specification of the method} \label{specification}
We now reconcile the general methodology with the biological rationale of the application. For this, we consider the estimators~\eqref{lassos} with $\ell_1$-penalization to exploit potential sparsity of the graphs. The \addinfo\ $D_{ij}$ is the Euclidian distance between brain regions $i, j$, and the link function is defined as $f(x)=x^3$ to capture the three-dimensionality of the coordinates. This  yields tuning parameters of the form $(\mathbf{r}^j)_i= \bar r^j \cdot D_{ij}^3$. We compare our approach \ourmethod\ to THR, MB, and GLASSO. Again, we adopt the ``and-rule'' for SI and MB. The tuning parameter in MB  as well as $\bar r^j$ in SI are selected via 10-fold cross-validation. Since the tuning parameter in GLASSO is not amenable to the same cross-validation scheme, GLASSO is calibrated via BIC. In absence of established selection criteria for thresholding, THR is calibrated such that the number of connections equals the number of connections for SI.

Note that the coordinates of the observations in the present data set correspond to volume regions in the brain. The observations in another type of fMRI-induced data that received considerable attention recently correspond to the cerebral cortex~\cite{cortex}. We would then recommend our approach with geodesic distances among the regions as \addinfo\ and a link function of the form $f(x)=x^2$ to capture the two-dimensionality of the spherical coordinates.



%% file: stable.tex
\newcommand{\tablereproducibility}{\begin{table}[t]
\centering
\caption{\ourmethod\ has higher reproducibility on the fMRI data set than standard methods}
\label{sample-table1}
\begin{center}
\begin{tabular}{l c  }
\hline
{\bf Method}  &{\bf Reproducibility (SD)} \\ 
\hline
\ourmethod        &62\ (7)\\ 
MB       &42 (6)\\
GLASSO        &55 (7)\\ 
\hline 
\end{tabular}
\end{center}
\end{table}}
 

\subsection{Reproducibility study} Reproducibility is a basic scientific principle. It requires that experiments and analyses can be replicated, and that these replications lead to the same scientific conclusions. This means in particular that the statistical methods involved need to be stable in the sense that two estimates should be commensurate if the corresponding data sets are collected in the same fashion.

An effective way to probe reproducibility of a method is data splitting. Indeed, agreement of estimates based on two parts of a data set under consideration indicates that the method is stable, while largely disagreeing estimates  raise a red flag. To compare the reproducibility of different methods on the fMRI data,  we proceed as follows. Given a patient and a method, the data describing the $210$~samples are split randomly into two sets of $105$ samples; then, reproducibility in terms of percentage agreement  $(\,1-{d_{\operatorname{H}}(\hat \edgeset, \tilde\edgeset)}/({|\hat \edgeset|+|\tilde\edgeset|})\,)\%$ is calculated for the graphs $\hat\edgeset$ and $\tilde\edgeset$ that are estimated on the two halves of the data. Averages are finally taken over $20$ random splits and over all the patients. 
 

\tablereproducibility

Table~\ref{sample-table1} shows that \ourmethod\ has the highest reproducibility among the method considered. (Note that the sample sizes in the split data are not sufficiently large for the inclusion of THR.) This suggests that the \addinfo\ is integrated effectively and that \ourmethod\  makes graph estimation on the data set under consideration more reliable. 


%% file: connect.tex
\newcommand{\plotallbrains}{
\begin{figure}[h]    
\includegraphics[width=1.2\textwidth,center]{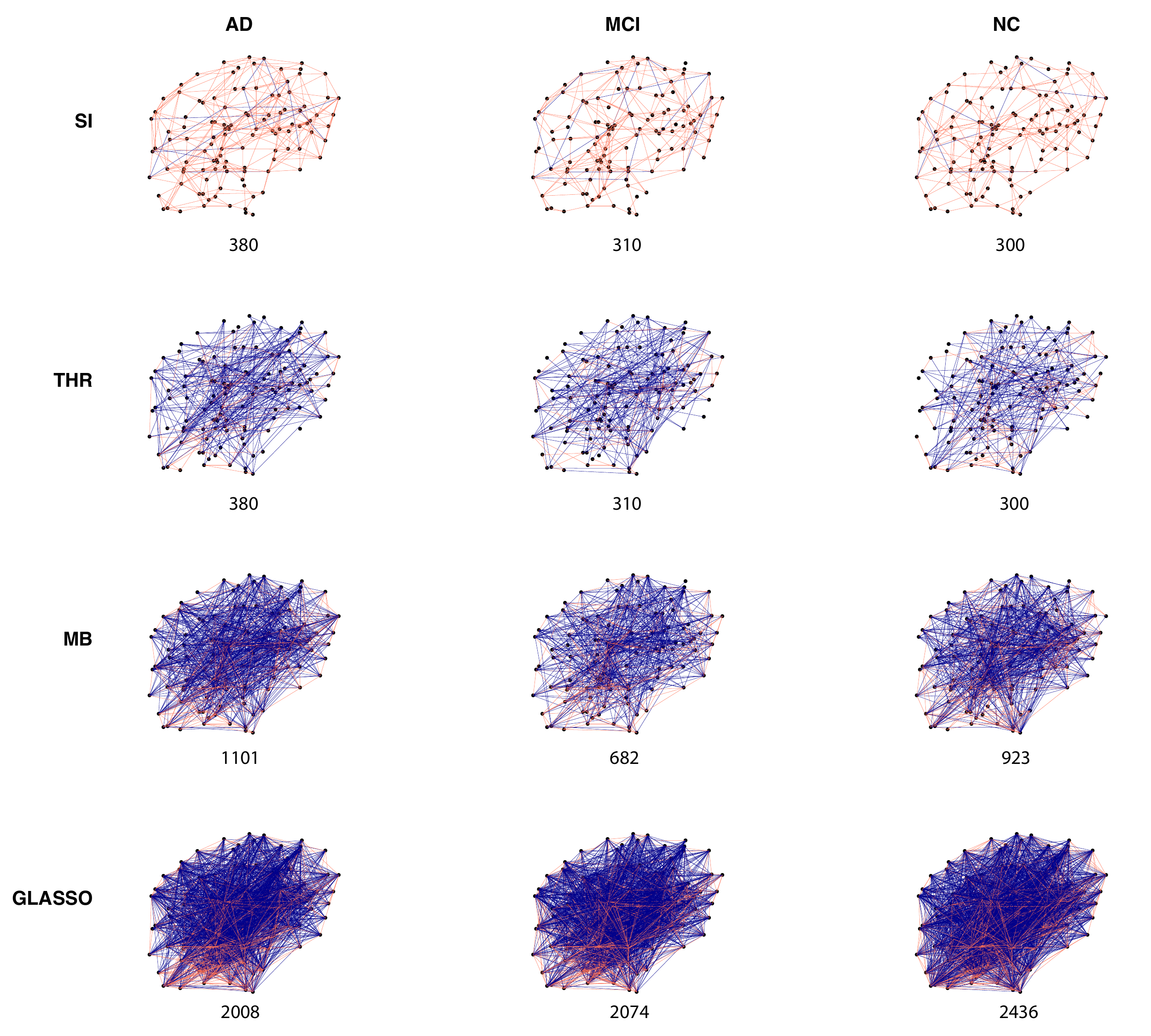}
\caption[]{Anatomical maps of the estimated brain connectivity networks show that in contrast to standard methods, \ourmethod\ entails direct connections mostly between spatially close regions (orange lines).}
\label{side}
\end{figure}
}

\plotallbrains 

\subsection{Connectivity remarks} After comparing the methods in terms of reproducibility, we give some brief remarks about the final graph estimates. 

To highlight differences among the methods, graphs for one subject from each of the study groups are displayed in Figure~\ref{side}. The graphs are estimated based on all the data available for one given subject at a time. Each column in the figure corresponds to one (fixed) subject of the three groups AD, MCI, and NC; each row corresponds to one of the four methods \ourmethod, THR, MB, and GLASSO. The plots show anatomical maps, that is, the nodes are positioned according to their 3D coordinates. Edges between near regions (distances in the lower quartile) are colored orange; edges between more remote regions are colored blue. The total number of edges is stated below each map. While showing only one set of graphs for illustration, we find  two patterns across all subjects: (i)~With cross-validation as the most common calibration scheme applied, \ourmethod\ yields the most sparse networks. This finding shows that \ourmethod\ can lead to more manageable  models. (ii)~\ourmethod\ yields a small number of edges between distant regions, but most edges are between spatially close regions. In strong contrast, the other methods do not exhibit such a preference. This finding confirms our expectations about the four methods, and it shows that the estimates provided by \ourmethod\  are in agreement with the biological rationale.

\begin{figure}[h!] 
    \textbf{AD-NC}\par\medskip
	\vspace{-5 mm}
	\hspace{-20 pt}
\includegraphics[width=1.5\textwidth, center]{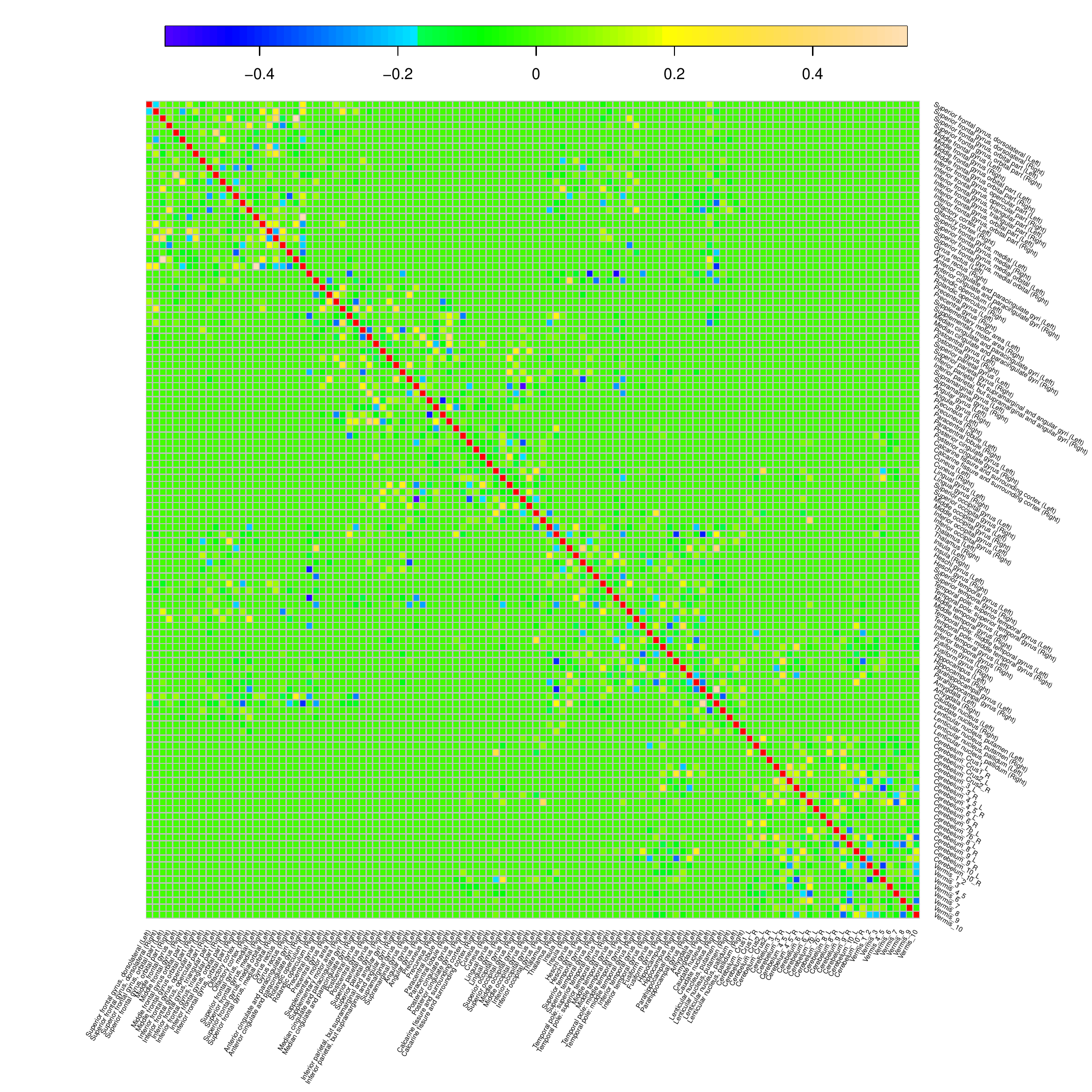}
\caption[]{Differences of average brain connectivity networks between the AD group and the NC group. Yellow indicates that an edge occurs more frequently in the AD group; blue indicates that an edge occurs more frequently in the NC group. The diagonal is colored in red. While the networks seem similar overall, there also seem to be considerably different connectivities between some spatically close regions.}
\label{AD-NC}
\end{figure}

To highlight differences among the study groups, the differences between the average graph estimates of \ourmethod\ for the AD and NC groups  are displayed in Figure~\ref{AD-NC}. Since accurate graph estimation is the basis for accurate group comparisons, we expect that \ourmethod\ can help uncovering differences and similarities among study groups in general. In the following, we briefly discuss some insights obtained for the data set at hand, leaving a complete biological study for future work. Note first that entries in the heat-map in Figure~\ref{AD-NC} denote the frequencies of the  edges in the AD group minus the corresponding frequencies in the NC group. The columns and rows are arranged such that spatially close regions tend to be close in the figure. We make two observations: (i)~The overall graphs for the two groups seem to be similar, but there also seem to be pronounced differences in the connectivities of some regions. This finding suggests that some connections might be affected by AD more than others. (ii)~Differences are found predominantly between spatially close regions (see blue and yellow entries close to the diagonal). This finding hints at the possibility  that AD concerns short-range direct connections more than long-range direct connections. (iii)~The plot contains areas with high variability (blocks with high concentrations of blue and yellow entries) such as the cerebelum close to the bottom right corner. The plot thus locates areas of interests for further studies.

%% file: diss.tex
We have shown that strengthening the role of tuning parameters is  an effective approach to incorporating \addinfo. In particular, we have shown that this approach can improve reproducibility, is computationally convenient, and has a clear Bayesian interpretation. 

Our scheme allows for the inclusion of application-specific information matrices~$D$ and link functions~$f$. The form of the information matrix~$D$ is typically predetermined in a given application. Similarly, the form of~$f$ often follows  clear scientific rationales as can be seen in our  brain connectivity example. One could also consider data-driven selections of~$f$ from a set of candidate functions by using score testing or related methods, cf.~\cite{scoretesting};  however, the sample sizes of typical data might be  too small for a thorough non-parametric selection of the link function. Importantly, via the link function~$f$ and the information matrix~$D$, our approach can be tailored to a wide range of application in bioinformatics, speech recognition, computer vision, and digital communications. For example, we envision further implementation in genomics, where the goal is to estimate gene regulatory networks based on gene expression levels~\cite{GRN}. The \addinfo\ that researchers commonly want to invoke for this are previously established sub-networks or networks estimated in other studies.

Finally, another feature of our concept is that it is readily amenable to theory and further extensions. We expect that sharp theoretical guarantees can be proved using standard techniques in the field~\cite{high}. Extensions to heavy-tailed or otherwise non-Gaussian data could be approached based on~\cite{nonGaussian}.  For the development of confidence sets for  graph estimation with \addinfo, we expect to combine our ideas with~\cite{Inference for graph}.